\pdfoutput=1
\documentclass[lettersize,journal]{IEEEtran}
\usepackage{algorithmic}
\usepackage{algorithm}
\usepackage{array}
\usepackage[caption=false,font=normalsize,labelfont=sf,textfont=sf]{subfig}
\usepackage{textcomp}
\usepackage{stfloats}
\usepackage{url}
\usepackage{verbatim}
\usepackage{graphicx}
\usepackage{cite}
\usepackage{amsmath,amssymb,amsfonts}
\usepackage[utf8]{inputenc}
\usepackage{csquotes}
\usepackage{booktabs}
\usepackage[capitalize]{cleveref}
\usepackage{diagbox}
\usepackage[table,xcdraw]{xcolor}
\usepackage{threeparttable} 
\usepackage{multirow}
\usepackage{lscape}
\usepackage{xcolor}
\usepackage{float}

\begin{document}

\title{A Framework for Semantics-based Situational Awareness during Mobile Robot Deployments}

\author{Tianshu Ruan, Aniketh Ramesh, Hao Wang, Alix Johnstone-Morfoisse, Gokcenur Altindal, Paul Norman, Grigoris Nikolaou, Rustam Stolkin and Manolis Chiou

\thanks{Tianshu Ruan, Aniketh Ramesh, Hao Wang and Rustam Stolkin are with the Extreme Robotics Lab (ERL) and National Center for Nuclear Robotics (NCNR), University of Birmingham, UK (e-mail: txr094@student.bham.ac.uk; anikethramesh94@gmail.com; hwang0721@outlook.com; r.stolkin@bham.ac.uk)}
\thanks{Alix Johnstone-Morfoisse, Gokcenur Altindal and Paul Norman are with Birmingham Centre for Nuclear Education \& Research (BCNER), University of Birmingham, UK (e-mail: amm960@student.bham.ac.uk; gxa992@student.bham.ac.uk; p.i.norman@bham.ac.uk)}
\thanks{Grigoris Nikolaou is with Department of Industrial Design and Production Engineering, University of West Attica, Greece (e-mail: nikolaou@uniwa.gr)}
\thanks{Manolis Chiou was with the Extreme Robotics Lab (ERL) and National Center for Nuclear Robotics (NCNR), University of Birmingham, UK, He is now with Queen Mary University of London (e-mail: m.chiou@qmul.ac.uk)}

}



\maketitle

\begin{abstract}
Deployment of robots into hazardous environments typically involves a ``Human-Robot Teaming'' (HRT) paradigm, in which a human supervisor interacts with a remotely operating robot inside the hazardous zone. Situational Awareness (SA) is vital for enabling HRT, to support navigation, planning, and decision-making. This paper explores issues of higher-level ``semantic'' information and understanding in SA. In semi-autonomous, or variable-autonomy paradigms, different types of semantic information may be important, in different ways, for both the human operator and an autonomous agent controlling the robot. We propose a generalizable framework for acquiring and combining multiple modalities of semantic-level SA during remote deployments of mobile robots. We demonstrate the framework with an example application of search and rescue (SAR) in disaster response robotics. We propose a set of ``environment semantic indicators" that can reflect a variety of different types of semantic information, e.g. indicators of risk, or signs of human activity, as the robot encounters different scenes. Based on these indicators, we propose a metric to describe the overall situation of the environment called ``Situational Semantic Richness (SSR)". This metric combines multiple semantic indicators to summarise the overall situation. The SSR indicates if an information-rich and complex situation has been encountered, which may require advanced reasoning for robots and humans and hence the attention of the expert human operator. The framework is tested on a Jackal robot in a mock-up disaster response environment. Experimental results demonstrate that the proposed semantic indicators are sensitive to changes in different modalities of semantic information in different scenes, and the SSR metric reflects overall semantic changes in the situations encountered.
\end{abstract}


\section{Introduction}
Situational Awareness (SA) is vital for robots deployed in the field to function with sufficient autonomy, resiliency, and robustness. This is especially true for Human-Robot Teams (HRT) in safety-critical applications such as disaster response, remote inspection in unstructured environments, or nuclear operations \cite{chiou2022robot,10018727,rustam2023status}.  In all cases, humans and robots require SA to make plans or decisions in the context of HRT (e.g. identifying a proper timing to switch control between human operators and robots). Hence, Humans and robots need to know and share what is happening in the environment to plan and act in a safe and coordinated manner. Humans and robots (to avoid verbosity we sometimes use the term ``robot'' synonymously with the AI or autonomous agents used to control the robot) have different and distinct strengths and weaknesses in terms of perception, ability to interpret sensory data, and ability to plan and execute decisions in response to that data in real time.

Building on low-level signals from multiple modalities of on-board cameras and sensors, higher-level ``semantic'' understanding \cite{10018727} of scenes, environments and situations must be developed. Often, this higher-level semantic knowledge will be critical for determining subsequent decisions and actions. Recent advances, especially from the computer vision community \cite{FCN, 3Dsemantic}, have begun to provide autonomous agents with some elements of semantic-level perception. Meanwhile, in real-world robotic systems at present, the intelligence of human operators may often be necessary to correctly interpret and act upon semantically rich situations. In this paper, we propose a framework for robots to acquire semantically enhanced SA that combines with human understanding in an explainable and intuitive way.

\begin{figure}[htbp]
    \centering
    \includegraphics[width=8cm]{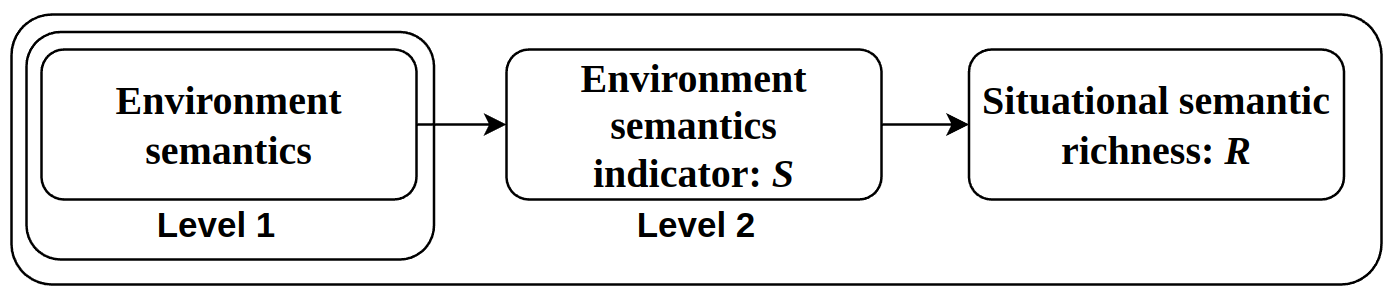}
    \caption{Mobile robot semantic SA.}
    \label{fig: robot semantic SA}
\end{figure}

Human factors SA can be modeled as three levels of awareness\cite{endsley2017toward}: Level 1) perception of elements in the current situation; Level 2) comprehension of the current situation; Level 3) projection of future status. In the robotics and AI research literature, it is common to use terms such as sensing, perception, scene understanding, semantics, and context \cite{bavle2021slam} instead of SA. There are connections among these related concepts, e.g. the concept of perception ``elements'' in SA can be linked to the ``semantics'' concept in AI. Hence, although the conventional SA model is designed to represent the awareness of human operators, the SA of an autonomous or semi-autonomous robot can be structured similarly in the scope of semantics. Elements of Level 1 SA can be objects, sensor readings, and other low-level semantics \cite{10018727}. The comprehension of the current situation at Level 2 corresponds to high-level semantics \cite{10018727}  (See \cref{fig: robot semantic SA}). Prediction, planning, or decision-making based on these are the main focus in Level 3.

Our work aims to build a systematic framework and concepts that: a) make SA sharing from robot to human easier, practical, and intuitive; and b) facilitate the use of semantically enhanced SA in HRT planning and decision-making frameworks. We build upon our previous work that proposed a taxonomy of semantic information \cite{10018727} and definitions of low-level semantics, high-level semantics, and the context in robot-assisted disaster response. 

In this paper, we explore Level 2 SA and propose the concept of an ``environment semantic indicator'' along with an example realization. Each indicator captures the understanding of an environment's semantics intensity (e.g. signs of human activity). Furthermore, we build upon those indicators to develop an aggregated metric, the ``Situational Semantics Richness'' (SSR), which expresses the overall intensity and plethora of semantic information in an environment. 

\section{Related work}
Early studies in human factors analyze SA based on human feedback after trials \cite{stanton2017state}, e.g. using the Situation Awareness Global Assessment Technique (SAGAT) \cite{endsley1988situation} or the Situational Awareness Rating Technique (SART) \cite{taylor2017situational}. Endsley and Mica R. proposed a three-level SA model, which is widely accepted \cite{endsley1995measurement}. SA is a subjective concept based on objective reflections of the environment, meaning everyone may understand the situation differently. Thus, building a generalized framework to regulate understanding is essential. A subjective scoring or weighting system that delivers subjective understanding is commonly used to differentiate each situation. For instance, the authors in \cite{hooey2011modeling}, build up a heuristic scoring system and give weights to different "situational elements" to model the SA obtained from aircraft pilots. \cite{mcaree2018quantifying} gives examples of formalizing some specific awareness, e.g. position and air environment (consisting of air traffic, airspace restrictions, and weather) using a scoring system. 

These works show how humans obtain SA. Elements of such approaches can be generalized to robot SA. Some researchers discuss human SA and robot SA combined or view the problem from a global perspective in an HRI context \cite{dini2017measurement}. 
Other researchers employ ontology to obtain the SA. Ontology concerns what kinds of things exist, how they can be organized, and what relationships exist between them \cite{huang2019ontology,tenorth2017representations}. \cite{armand2014ontology} models simple situations on the road and crossroads using ontology. Authors categorize road contexts into "mobile entities", "static entities", and "context parameters" that describe the relationship between entities from the spatio-temporal scope. Rules are established for the vehicle when the combination of road contexts changes. Ontologies are intuitively straightforward for modeling situations and are easy to understand. However, the ontology models are built on simplified or specific situations. They may have problems in complex environments and unexpected situations. Hence, robots need multiple inference methods to obtain SA \cite{tenorth2017representations}. Alternatively, probabilistic methods can be used to model the environment and generate SA \cite{shuang2014quantitative}. In \cite{nguyen2019review}, authors compare multiple SA measurements that formalize the SA. Apart from a human perspective, the authors also review the SA for Unmanned Aerial Vehicles (UAV). They claim that most SA studies focus on the human perspective and indicate there are limited methods to frame and obtain UAV SA. 

In general, most of these SA assessments define metrics highlighting the flexibility and the importance of expert knowledge. However, there are limited works on how robots perceive high-level semantics and how robots can aggregate those semantics into coherent and usable metrics reflecting the overall SA and context. Unlike human SA research, most robot SA research still focuses on addressing specific problems from one specific scope, e.g. electromagnetic jamming security \cite{gao2020uav}, or failure conditions \cite{ginesi2020autonomous,ghezala2014rsaw}. In contrast, we propose a general framework and an example realization for an aggregated metric of SA which enables robots to understand the overall environment situation and can be generalized to different deployment tasks.

\section{Problem Formulation and Concepts Definition}
Here, we assume that the deployed robots need to perform tasks such as scanning a damaged building \cite{kruijff2012rescue}, surveying and sampling contamination in a hazardous site \cite{nagatani2013emergency}, remotely inspecting and monitoring facilities, or searching for human victims \cite{murphy2014disaster, 10018727} in the context of disaster response. Robots can be teleoperated \cite{chiou2022robot}, semi-autonomous (e.g. variable autonomy \cite{reinmund2024variable, methnani2024s}, mixed-initiative \cite{chiou2021mixed} or shared control \cite{pappas2020vfh+} paradigms) or run fully autonomously. In all cases, robots need SA to make plans or decisions $\mathcal D$ in the context of HRT. Hence, there is a mapping $I:\mathcal S \mapsto \mathcal D$ between a set of environment semantics $ \mathcal S$ and the decisions $\mathcal D$. 
  
Specifically, $\mathcal S=\{S_1, S_2, ..., S_i, ...S_n\}, (n >=1, 0<S_i<1)$ comprises a set of different possible types of environment semantics, e.g. Signs of Human Activities (SHA), noise for LiDAR, or detection of hazards, where $n$ denotes the number of semantic indicators in $\mathcal S$. Note that $\mathcal S$ can be configured to contain many different kinds of semantic information, as may be appropriate to different types of robot missions and application domains. Without loss of generality, in this paper we use the example of disaster response, to provide an intuitive illustration of how this framework can be applied in a practical task. As examples of possible $\mathcal S$, we present experiments in which we use: $S_1 = S(noise)$, $S_2 = S(risk)$, $S_3 = S(SHA)$, and $S_4 = S(radiation)$. These are examples of environment semantics that can be useful in disaster response and remote inspection missions \cite{10018727}.

A significant challenge is that it is non-trivial to parameterize a framework for a mapping $I$, which can \textit{directly} map semantics $\mathcal S$ onto decisions $\mathcal D$. Therefore, the key idea of this paper is to introduce an intermediary term, which we call Situational Semantics Richness (SSR) $R$. The term $R$ serves to aggregate the environment semantics combinations in $\mathcal S$, which can then assist with bridging towards the decision set $\mathcal D$. We define the function $R=f(\mathcal S_i, W_i)$, where $W_i$ are a set of weights that reflect the relative impact of each type of semantic information. Note that this paper is focused on addressing the problem of progressing from $\mathcal S$ to the intermediary term $R$. The next challenge, of formulating a relationship between $R$ and $\mathcal D$ will form the subject of a future paper and is out of the scope of the present paper. However, in this paper, we show how the formulation of $\mathcal S$, and its mapping to intermediary term $R$, is already a useful tool in its own right, for assisting SA in HRT missions.

\section{Environment semantic indicator} \label{semantics}
\subsection{Laser noise intensity}
Many Unmanned Ground Vehicles (UGV) rely on lasers for autonomous navigation. However, laser noise potentially affects the navigation. We adapt the method to obtain laser noise variance ($\sigma^2_{noise}$) in our previous work \cite{ramesh2022robot}. It is calculated by convolving the laser map image with a 3x3 mask and applying summations on the resultant matrix. Then, we adapt the noise variance into a sigmoid function (See \cref{eq:laser noise} and \cite{ramesh2022robot}) to obtain the laser noise intensity $S(noise)$ The noise variance thresholds ($\sigma^2_{noise} \approx 1.4$) is obtained by our test in the environment of halting the robot navigation based on our hardware setup. Combining the preliminary test results, the laser noise intensity is designed as: 

\begin{equation}
\begin{aligned}
S(noise) = \frac{1}{1+\exp(-a\cdot \sigma^2_{noise}+a\cdot b)}
\\
\label{eq:laser noise}
\end{aligned}
\end{equation}
where $a = 5$ and $b = 1$ to have a curve that responds to medium inputs but is not oversensitive to low and high inputs

\subsection{Risk to robots}

Risk to robots can be quantified based on hazard level and hazard length \cite{soltani2004fuzzy}. Hazard level refers to how dangerous the hazard is to the robot, and hazard distance refers to the distance to the object. Furthermore, time may also affect the risk level, e.g. the accumulated dose received from the radiation sources.

In this work, we assume that the robot can detect potential risks by detecting HAZMAT signs that commonly exist in hazardous environments. In general, humans and robots face similar risks. However, considering the slight difference between risk to humans and risk to robots, the categorization might differ but can be trivially adapted to reflect human risks, expert knowledge (e.g. by first responders), or different scenarios. We make a categorization of HAZMAT signs into three levels heuristically: low risk or no risk for the robot (e.g. poison, infectious substance, nonflammable gas, inhalation hazard), medium risk (delayed hazard to a robot or they can be high risks under certain occasions, e.g. corrosive, radioactive, dangerous when wet, oxygen, organic peroxide), and high risk (immediate hazard to the robot, e.g. explosives, flammable solid, flammable gas, spontaneously combustible material). Other risks not included can also be added trivially in different scenarios. 

Intuitively, distance is a factor that relates to the risk intensity. Referring to the relationship between radiation strength and distance \cite{voudoukis2017inverse}, we apply a similar model to risk by using the inverse square law:

\begin{equation}
\begin{aligned}
S(risk) = \sum_{i=1}^{n}\frac{H_{j}}{d_{i}^2}
\label{eq:risk}
\end{aligned}
\end{equation}
, where $n$ is the number of detected signs; $i$ is the label of each detected sign; $d$ refers to the distance to the robot; $j$ refers to the level of the corresponding signs; $H_{j}$ refers to the risk intensity of each level of HAZMATs sign.

Then, we normalize the risk score by applying the sigmoid function. The reason for using the sigmoid function is when the x-axis closes to infinity, the slope of risk is low and accords with human common understanding i.e. the environment which has 6 high-risk objects has a similar $S(risk)$ with the environment containing 5 high-risk objects. The normalized score is:

\begin{equation}
\begin{aligned}
S(risk)_{norm} = \frac{1}{1+e^{-aS(risk)-b}}
\label{eq:risk_norm}
\end{aligned}
\end{equation}
, where $a$ and $b$ are used for tuning the functions. In our experiments, we set them as $a=0.09$ and $b=-4.8$ heuristically to get a meaningful and usable curve. Experts can tune $a$ and $b$ for different tasks.

\subsection{Sign of human activity}
Robots might not always directly detect human victims in the environment (e.g. trapped under debris or occluded by objects). Hence, robots must identify clues to find victims. Signs of Human Activity (SHA) are considered a potential factor in finding people \cite{yangvisual}. We use human belongings, e.g. mobile phones, keys, and watches, as indicators of human activity. Intuitively, the dispersion of personal belongings makes a difference to the SHA. Thus, the SHA is modeled from two aspects: class of objects and dispersion of the objects.

We design three classes of objects to differentiate the impact of different human belongings: i) high impact means there is a high chance to be found on the human body (e.g. glasses, key, cell phone, watch); ii) medium impact means there are chances to be found in proximity to the human body (e.g. cap, mask, wallet); iii) low impact means there is a high chance not be carried on the human body (e.g. laptop, backpack). This heuristic classification is an example of realization for our framework, and experts can adjust it.

Based on the above, we propose the following model to estimate the sign of human activity score $S(SHA)$:

\begin{equation}
\begin{aligned}
    dispersion = \sum_{i=1}^{n}|d_{i}-d_{average}|
    \\
    S(SHA) = \sum_{i=1}^{n}\frac {P_{j}}{dispersion}
    \\
    \label{eq:SHA}
\end{aligned}
\end{equation}

where $n$ is the number of detected objects, $i$ is the label for each object, $d_{i}$ is the distance from the corresponding objects to the robot, $d_{average}$ is the average distance of all the objects, $j$ is the label of class for objects, $P_{j}$ refers to the impact of corresponding objects.

Similarly, we normalize the $S(SHA)$ using the sigmoid function:
\begin{figure*}[htbp]
    \centering
    \includegraphics[width=16cm]{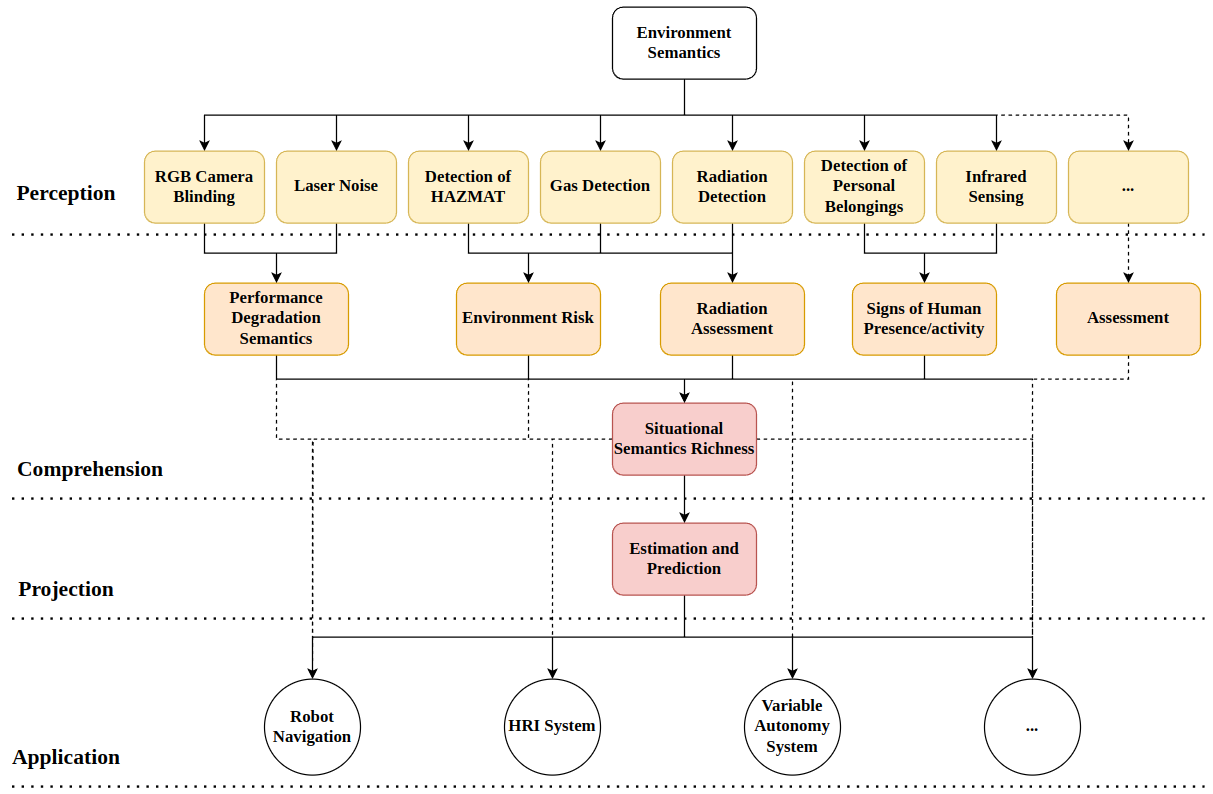}
    \caption{Semantics-based SA framework: yellow box refers to low-level semantics, orange box refers to high-level semantics and red box refers to context. Black dash lines indicate the potential connections among different levels.}
    \label{fig: environment semantics}
\end{figure*}

\begin{equation}
\begin{aligned}
S(SHA)_{norm} = \frac{1}{1+e^{-a(SHA)-b}}
\label{eq:SHA_norm}
\end{aligned}
\end{equation}

where $a$ and $b$ are used for tuning the function. In our experiments, we used $a=10$ and $b=-0.5$ to have a curve that has similar sensitivity characteristics with \cref{eq:laser noise}. The parameters can be adjusted when expert knowledge is involved.

\subsection{Radiation}

Not only humans but robots may also be affected by radiation \cite{nagatani2013emergency}. The highest risk will be that of the onboard electronics, as radiation can cause disruptions, malfunction, or even complete failure of electronic components. These issues can be addressed by applying appropriate shielding and mitigation techniques or designing radiation-hardened robots. Alternatively, we might monitor the radiation strength in deployment so that robots and humans can avoid exposure.

The risk associated with radiation depends on the distance from the source and the radiation’s type, strength, and energy. It is important to distinguish between the dose rate and the total integrated dose. Dose rate is commonly measured in Sieverts per hour  ($Sv/h$), which is the rate at which the radiation is received at a given moment. The total integrated dose measured in Sieverts which results from the accumulation of radiation over time.

We designed a mapping $I:\mathcal G \mapsto S(radiation)$. It translates the raw readings which require radiation backgrounds for understanding to a matric ranging from 0 to 1. $\mathcal G$ is a set of gamma radiation dose rates $G$ in $\mu Sv/h$ from the sensor and $S(radiation)$ denotes the radiation strength. Specifically, the sigmoid function is applied in the mapping $I$ to normalize the $S(radiation)$:

\begin{equation}
\begin{aligned}
S(radiation)_{norm} = \frac{1}{1+e^{-aG-b}}
\label{eq:radiation_norm}
\end{aligned}
\end{equation}
where $a=8$ and $b=-8$. Specifically, $S(radiation)=0$ refers to no radiation, and $S(radiation)=1$ refers to the radiation that can instantly damage the robot. This tuning setting takes into account the effects of background radiation (under 0.4 seivert/h usually) and decrease the impact of it.

\section{Situational Semantics Richness} \label{richness}

With these semantics in \cref{semantics}, we propose a framework (Shown in \cref{fig: environment semantics}) that fuses semantics of the environment on a higher level, i.e. a metric that describes the overall status of the environment in an aggregate representation.

In real-world situations, if we do not have a dataset, applying data-driven approaches is impractical or intractable. To the best of our knowledge, no dataset involves all the environment semantics, i.e. it is not feasible to build a parametric model and train an end-to-end network to assess the situation. Moreover, SA is a subjective understanding and needs to be intuitive and explainable, especially in safety-critical and hazardous applications. Thus, we must capture human understanding into our framework. It is common practice to build a heuristic-based system that comprises important factors and expert knowledge to obtain SA. It is straightforward to adjust. The Situational Semantics Richness (SSR) is proposed based on this idea. We obtain the score of each environment semantic indicator and adapt them into the proposed framework by developing the SSR score, which expresses the overall intensity and plethora of semantic information from the environment.

We obtain a set of normalized metrics $S$ from 0 to 1 in \cref{semantics}. In different scenarios, different semantics might have different importance. To address this, we assign an importance weight $W_{i}$ for each $S_i$. Tuning $W_{i}$ gives the framework extensibility to different deployment cases and tasks while enabling leveraging expert knowledge. In our experiment, we adopt the exponential weight $W_{i} = {e}^\frac{S_{i}}{0.99}$ to emphasize the environment semantics with a higher score, where $S_{i}$ is the score of each environment semantic, $i$ is the label of environment semantics. The exponential weight lets high-score semantics impact more in the final SSR score. Then, we define the situational semantics richness $R$ as:

\begin{equation}
\begin{aligned}
R = \sum_{i=1}^{n} W_{i}\times S_{i}
\\
\label{eq:richness}
\end{aligned}
\end{equation}
where $n$ denotes the number of environment semantics. We normalize the $R$ using a sigmoid function to obtain $R_{norm}$, which can enable a better understanding of the SSR intensity by humans:

\begin{equation}
\begin{aligned}
R_{norm} = \frac{1}{1+e^{-aR-b}}
\\
\label{eq:richness_norm}
\end{aligned}
\end{equation}
where $a$ and $b$ can be set as 10 and -0.5 correspondingly to fit the range from (0,1).

To address the effects of unreliable scores caused by noise or false detection, we process the normalized SSR score by involving historical data. We apply attention mechanism regression to comprise the past SSR scores and emphasize the impact of the latest score. The attention mechanism was first proposed by Èlizbar A. Nadaraya \cite{nadaraya1964estimating} and Geoffrey S. Watson \cite{watson1964smooth}, and has been widely used for non-parametric estimation and deep learning \cite{vaswani2017attention}. It runs like the human's attention to indicate which value or factor deserves more focus among the rest of the data.

In our case, we can obtain a set of $R$ in time sequence $\mathcal R_t=\{R_1, R_2,..., R_t\}, t>0$, $t$ refers to the current timestamp. Then, the estimated situational semantics richness $\hat{R_t}$ at time $t$ can be defined as:

\begin{equation}
\begin{aligned}
\hat{R_{t}} = \sum_{i=1}^{n} \frac {K(t-t_{i})}{\sum_{j=1}^{n}K(t-t_{j})} \times \mathcal R_{t}
\label{eq:attention}
\end{aligned}
\end{equation}
, where $K$ is the kernel function, $t_i$ is the timestamp. If we apply the Gaussian kernel which is mostly used in \cref{eq: estimated attention}, the estimate situational semantics richness $\hat{R_t}$ is:
\begin{equation}
\begin{aligned}
K(u) = \frac{1}{\sqrt{2\pi}} \exp (-\frac{u^2}{2})
\\
\hat{R_{t}} = \sum_{i=1}^{n} softmax(-\frac{1}{2}(t-t_{i}^2)) \times \mathcal R_{t}
\label{eq: estimated attention}
\end{aligned}
\end{equation}
where $u$ refers to $t-t_i$ in this case.

In the experiment, we selected the time window of 5 latest $R_{t}$, which means $n=5$. According to the features of the applied attention mechanism, the older of the sample, the less impacts to the final score. Hence, 5 latest sample is enough to refer. Considering the sampling rate limitation from the radiation sensor (1 Hz), the updating rate of the SSR score is synchronized to 1 Hz. Hence, the tuned situational semantic richness $\hat{R_{t}}$ refers to the $R_{t}$ in 5 seconds and we apply the $\hat{R_{t}}$ as the final SSR score. The time complexity of the whole process is $O(n)$ and the space complexity is $O(n)$ as well, which means that it is an efficient algorithm in the scope of computation.  

\section{Experiments}
\label{Experiments}

We tested our framework intending to evaluate: i) if the framework can accurately perceive each semantics and their changes separately, and ii) if the framework is robust and can adapt in an environment with multiple levels and types of semantic indicators. 

We used a Jackal mobile robot with an Intel I5 CPU and GTX 1650TI GPU onboard. The framework is built based on the ROS Noetic system. We ran the framework directly on the Jackal to avoid image transferring to the offsite computer. Additionally, sensors, including a real-sense D435i camera, Velodyne vlp-16 Lidar, and Hamamatsu Gamma Sensor C12137 are mounted on the Jackal. We applied the Yolact \cite{yolact-iccv2019} as our vision model, providing object detection and instance segmentation results. We make some modifications to the system to enable us to attach depth data to each detected object by aligning the RGB image and depth image. The Hamamatsu Gamma Sensor C12137 is specifically designed to measure gamma radiation in the range 0.03 to 2 MeV, and dose rate up to 100 $\mu Sv/h$. Even though the robot would be able to detect high-strength radiation sources from a distance, constrained by regulations from the university sadly, we had to use a low-strength radiation source (uranium rock) that cannot be detected from long distances (over 10 centimeters). Stronger sources are not available in our project.

We assume the robot has a prior map from the SLAM, but SLAM is outside the scope of this paper. The proposed environment semantics were placed in the environment after the mapping. Then, we predefined a set of waypoints that the robot has to navigate. The reason for applying autonomous navigation is that we would like to keep the similar trajectories of the robot in corresponding trials.

\begin{figure*}[htbp]
\centerline{\includegraphics[width=16cm]{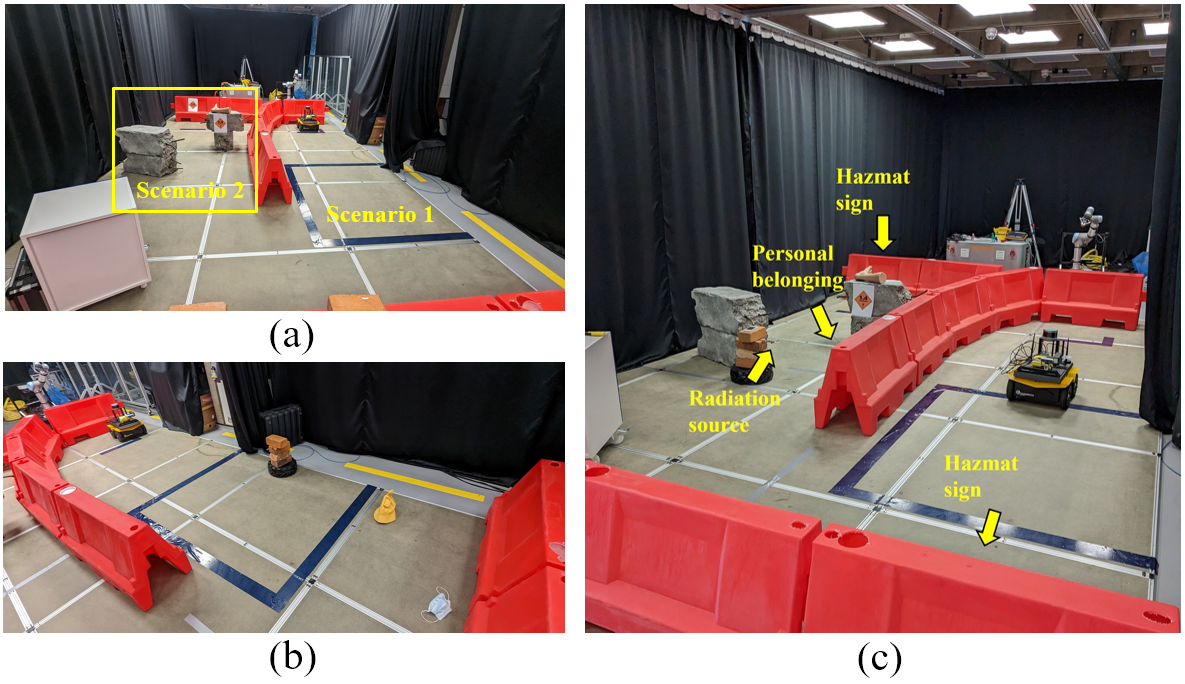}}
\caption{(a): Layout for Experiment I. The dark blue box area (scenario 1) on the ground is used for laser noise or the radiation source (uranium rock). The yellow box area (scenario 2) is used for hazmat signs or personal belongings; (b): Layout for two environment semantics scenarios of experiment II; (c): Layout for three environment semantics scenarios of experiment II. In the picture, some environment semantics are covered by red barriers.}
\label{fig:experimentI_layout}
\end{figure*}

\subsection{Experiment I}
\subsubsection{Implementation}
We tested the framework in the scenarios with single environment semantics in experiment I. We set two scenarios separately in the area (See \cref{fig:experimentI_layout} (a)). In each scenario, only one environment semantics was added i.e. each environment semantic is independent, and only one semantics with a big impact can be perceived at any time.

Specifically, we defined three cases to differentiate the intensity of the environment semantics: LOW (radiation and noise), MEDIUM (radiation and SHA), and HIGH (risk and noise), which correspond to the levels of environment semantics. For instance, the HIGH case refers to risk and noise that can be detected at a high level; the MEDIUM case refers to radiation and SHA detected is medium level. Due to the nature of the uranium rock, the high-strength radiation scenario needed teleoperation and positioning of the robot close to mock the situations in which the sensor receives a high dose rate. And we applied longer distances to mock MEDIUM and LOW cases. We ran the robots 10 times in three cases each.

\begin{figure*}
	\subfloat[SSR and environment semantics (radiation and noise) intensity timeline in LOW case.]{
		\begin{minipage}[t]{1\linewidth}
			\includegraphics[width=16cm]{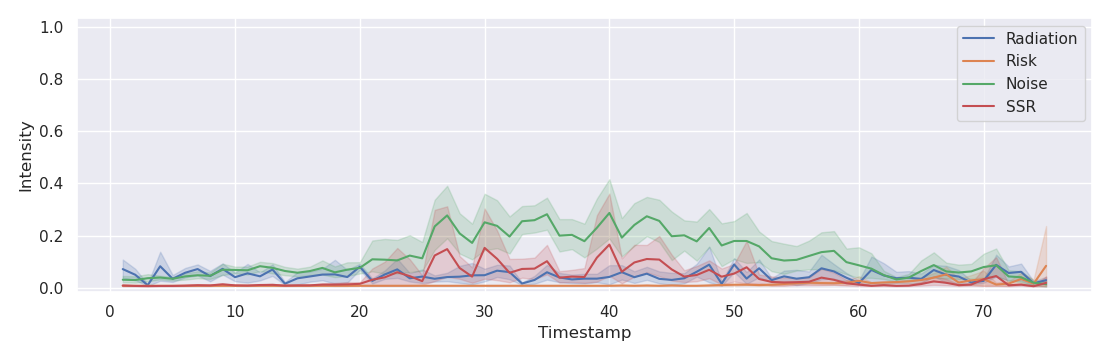}\\
			\vspace{0.02cm}
		\end{minipage}%
	}%
 
  \subfloat[SSR and environment semantics (radiation and SHA) intensity timeline in MEDIUM case.]{
		\begin{minipage}[t]{1\linewidth}
			\includegraphics[width=16cm]{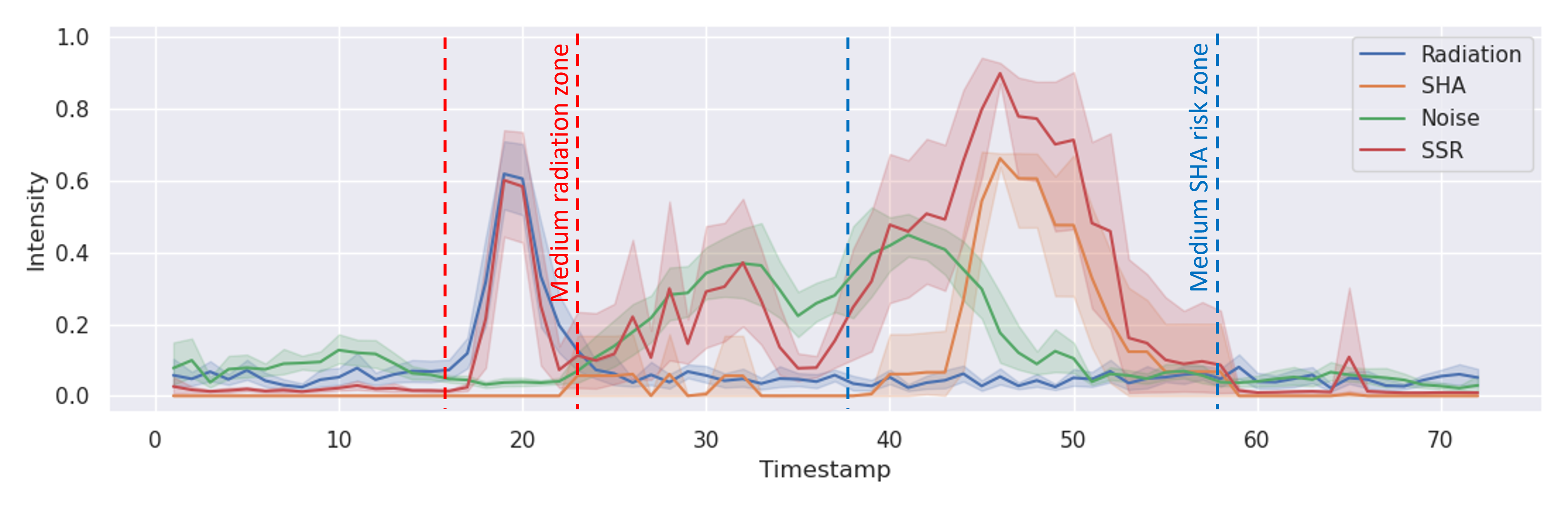}\\
			\vspace{0.02cm}
		\end{minipage}%
	}%
 
  \subfloat[SSR and environment semantics (risk and noise) intensity timeline in HIGH case.]{
		\begin{minipage}[t]{1\linewidth}
            \includegraphics[width=16cm]{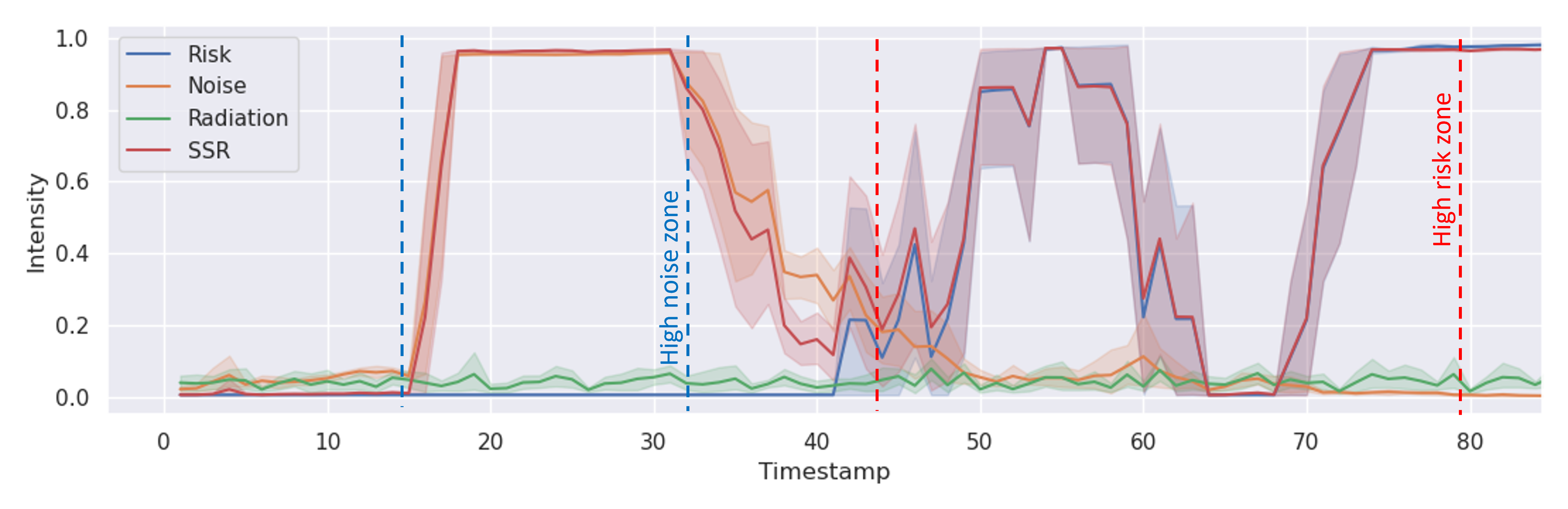}\\
			\vspace{0.02cm}
		\end{minipage}%
	}%
	\caption{SSR and environment semantics intensity timeline. Zones between dash lines refer to the corresponding semantics detected from the environment.}
	\vspace{-0.2cm}
	\label{fig:seperate_timeline}
\end{figure*}

\begin{table}
    \normalsize
    \centering
    \caption{Sensitive analysis ($S1$, $St$) of environment semantics with the SSR score in Experiment I}
    \begin{tabular}{c<{\centering}c<{\centering}c<{\centering}c<{\centering}c<{\centering}c<{\centering}}
        \toprule
        Case&Semantic Indicator&$S1$&$St$\\
        \midrule
        MEDIUM&Radiation (red zone)&0.997&0.997\\

         &SHA (blue zone)&0.977&0.978&\\

        HIGH&Noise (blue zone)&0.997&0.997\\

         &Risk (red zone)&0.766&0.769&\\
        
        \bottomrule
    \end{tabular}
    \label{table: Sensitive}
 \end{table}

\subsubsection{Results}

We collected data, including the scores of all environment semantic indicators and timestamps. To examine if the framework can differentiate environments with different levels of semantics, we generated the timeline of the SSR score and environment semantics in \cref{fig:seperate_timeline}. Note that, in \cref{fig:seperate_timeline} (c), there is a deep flat between 60-70 seconds. It was caused by the occasion that the robot turns momentarily and faces the black curtain on the left. At that point, the camera, which is constrained by the view of the field, was unable to see the last HAZMAT sign on the red barrier until it moved forward and faced the sign. To the best of our knowledge, there is no prior baseline or dataset to compare with. The figures from the different cases reveal that the framework is capable of correctly outputting the corresponding SSR scores in the individual environment semantics scenario. When the robot reached the scenario, the environment semantics indicator responded in time and the SSR score was more affected by the semantics with higher scores as designed i.e. the SSR score tracked the environment semantics with the highest score and magnitude. When zooming in specific environment semantics zones, we notice that low-intensity semantics do not affect the SSR score much if medium or high intensity semantics exist. We apply variance-based sensitivity analysis \cite{sobol2001global} in the cases MEDIUM and HIGH (See \cref{fig:seperate_timeline} (b) (c)). Specifically, we calculate the global sensitivity (first-order index $S1$, total-effect index $S_t$) of each environment semantics indicator to the SSR score in \cref{table: Sensitive}. We skip the LOW case to save space, as LOW case is not that important in real deployments as long as no significant error has been found. The analysis showed similar $S1$ and $S_t$ results. The individual semantics of the corresponding zones show strong sensitivity (over 0.7) to the SSR score. It indicates that the SSR score can respond quickly and accurately to single semantics changes in the environment. The medium or high intensity semantics dominates the impact and leads the changes in the SSR score. Our framework is tested to capture the changes in individual semantics and reflect correctly on the SSR score. 

\subsection{Experiment II}
\subsubsection{Implementation}
We tested the framework's performance in scenarios with concurrent multi-environment semantics in Experiment II. We designed 12 scenarios in this experiment (See \cref{table:12 scenarios}). Specifically, we pick all the possible two-semantics combinations with high and medium level intensity and two three-semantics combinations. The scenarios covered a wide spectrum of situations that robots may encounter. We did not test with all of the semantics concurrently as we were constrained by our vision system (train the HAZMAT and personal belonging detection separately). Additionally, the noise generation design \cite{RameshBraunRothfuss2023_1000159207} has only two levels of intensity available. High refers to deliberately adding laser noise into the scenario. Medium refers to no artificial laser noise added (e.g. normal noise caused by turning the robot). Low refers to background noise. In the scenarios with two environment semantics, we used the corresponding area in \cref{fig:experimentI_layout} (b). The dark blue box on the ground was used for laser noise or the radiation source. Hazmat signs were put on the right red barrier and personal belongings were scattered in front of the right red barrier. In the scenarios with three environment semantics, we applied laser noise in the dark blue box, hazmat signs, and radiation source as \cref{fig:experimentI_layout} (c) shows, and personal belongings are scattered around the gray bricks. The robot ran 5 times in each scenario.

\begin{table}
    \normalsize
    \centering
    \caption{Combinations of environment semantics in different scenarios in Experiment II}
    \begin{tabular}{c<{\centering}c<{\centering}c<{\centering}c<{\centering}c<{\centering}c<{\centering}c<{\centering}}
        \toprule
        Scenario&Radiation&Risk&SHA&Noise\\
        \midrule
        1&High&/&/&High&\\

        2&Medium&/&/&Medium&\\

        3&High&High&/&/&\\

        4&Medium&Medium&/&/&\\

        5&/&High&/&High&\\

        6&/&Medium&/&Medium&\\

        7&/&/&High&High&\\

        8&/&/&Medium&Medium&\\

        9&High&/&High&/&\\

        10&Medium&/&Medium&/&\\

        11&High&/&High&High&\\

        12&Medium&Medium&/&Medium&\\
        
        \bottomrule
    \end{tabular}
    \label{table:12 scenarios}
 \end{table}
 \begin{table}
    \normalsize
    \centering
    \caption{Spearman's rank correlations coefficient ($\rho$) of environment semantics with the SSR score in Experiment II}
    \label{table: correlations}
    \begin{tabular}{c<{\centering}c<{\centering}c<{\centering}c<{\centering}c<{\centering}c<{\centering}c<{\centering}}
        \toprule
        Scenario&Radiation&Risk&SHA&Noise\\
        \midrule
        1&0.747&/&/&0.996&\\

        2&0.863&/&/&0.220&\\

        3&0.669&0.798&/&/&\\

        4&0.563&0.880&/&/&\\

        5&/&0.731&/&0.666&\\

        6&/&0.477&/&0.755&\\

        7&/&/&0.825&0.785&\\

        8&/&/&0.782&0.455&\\

        9&0.679&/&0.865&/&\\

        10&0.667&/&0.794&/&\\

        11&0.533&/&0.442&0.907&\\

        12&0.424&0.627&/&0.498&\\
        
        \bottomrule
    \end{tabular}
 \end{table}

 \begin{table}
    \normalsize
    \centering
    \caption{Kendall's rank correlation coefficient ($\tau$) of environment semantics with the SSR score in Experiment II}
    \label{table: kendallcorrelations}
    \begin{tabular}{c<{\centering}c<{\centering}c<{\centering}c<{\centering}c<{\centering}c<{\centering}c<{\centering}}
        \toprule
        Scenario&Radiation&Risk&SHA&Noise\\
        \midrule
        1&0.618&/&/&0.598&\\

        2&0.702&/&/&0.182&\\

        3&0.523&0.567&/&/&\\

        4&0.426&0.731&/&/&\\

        5&/&0.529&/&0.509&\\

        6&/&0.420&/&0.517&\\

        7&/&/&0.671&0.567&\\

        8&/&/&0.633&0.316&\\

        9&0.548&/&0.698&/&\\

        10&0.519&/&0.626&/&\\

        11&0.389&/&0.337&0.742&\\

        12&0.311&0.517&/&0.397&\\
        
        \bottomrule
    \end{tabular}
 \end{table}

\subsubsection{Results}
\begin{figure*}
    \centering
    \subfloat[Radiation-noise combination]{
        \includegraphics[width=0.48\linewidth]{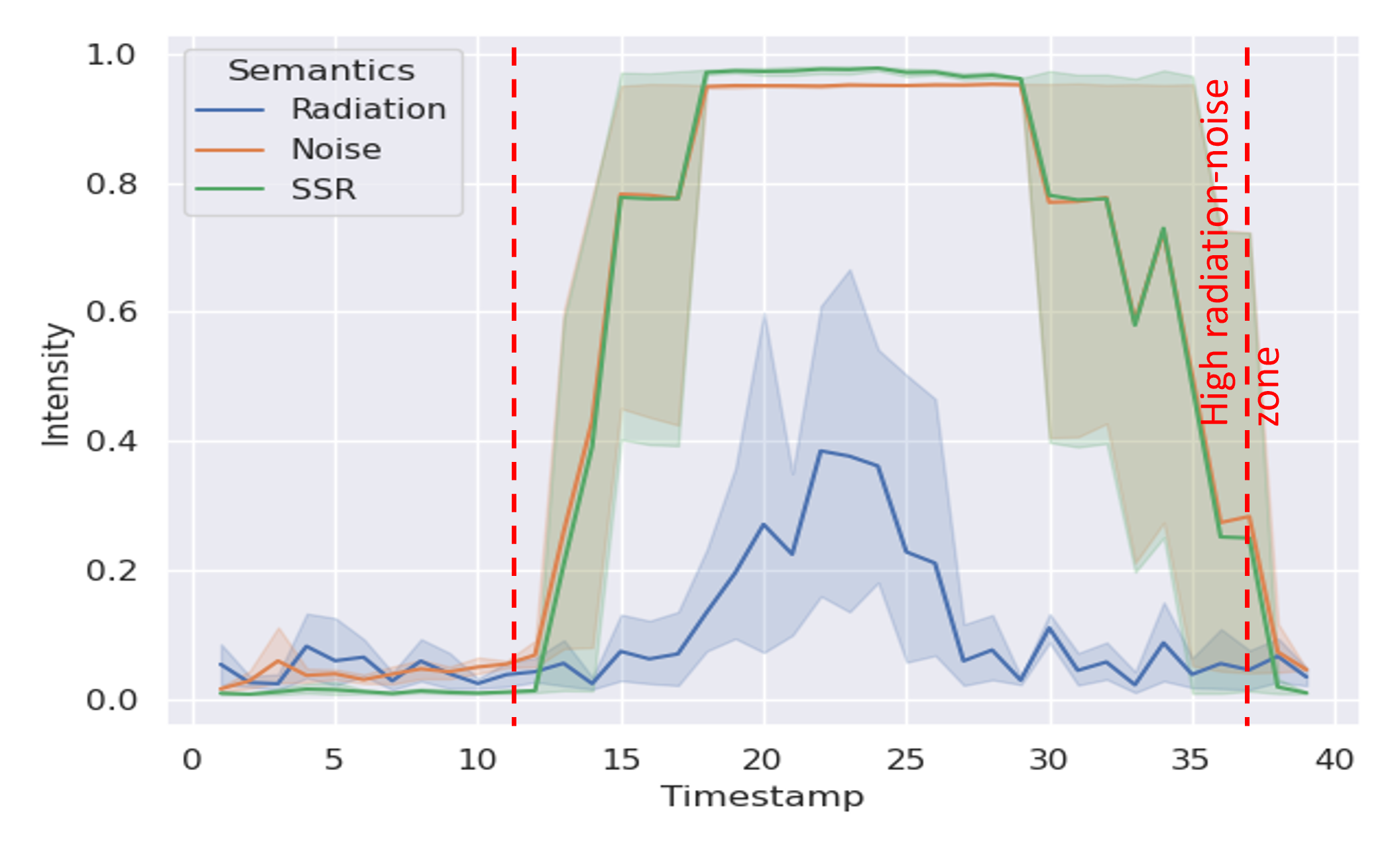}
        \includegraphics[width=0.48\linewidth]{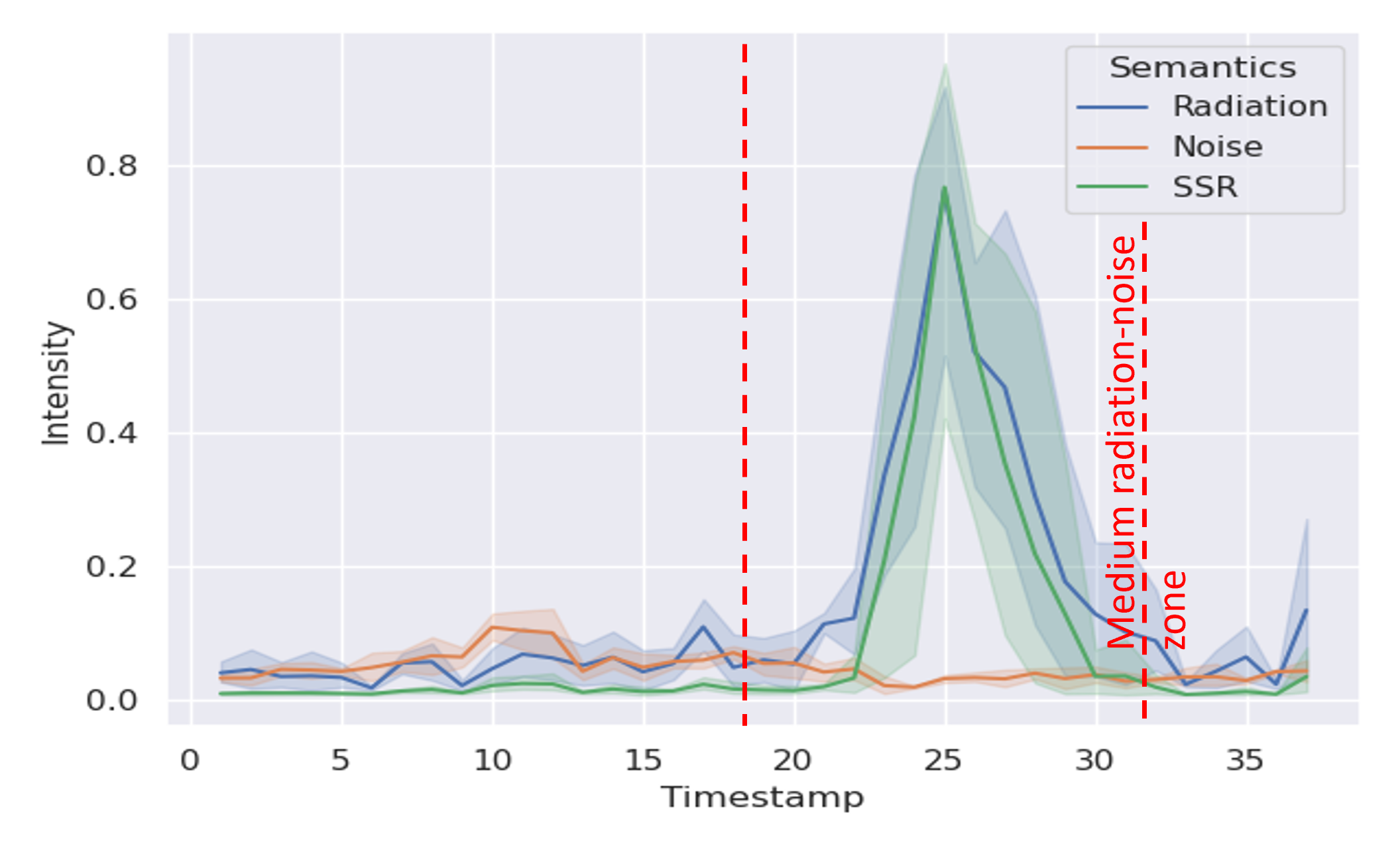}
    }
    
    \subfloat[Radiation-risk combination]{
        \includegraphics[width=0.48\linewidth]{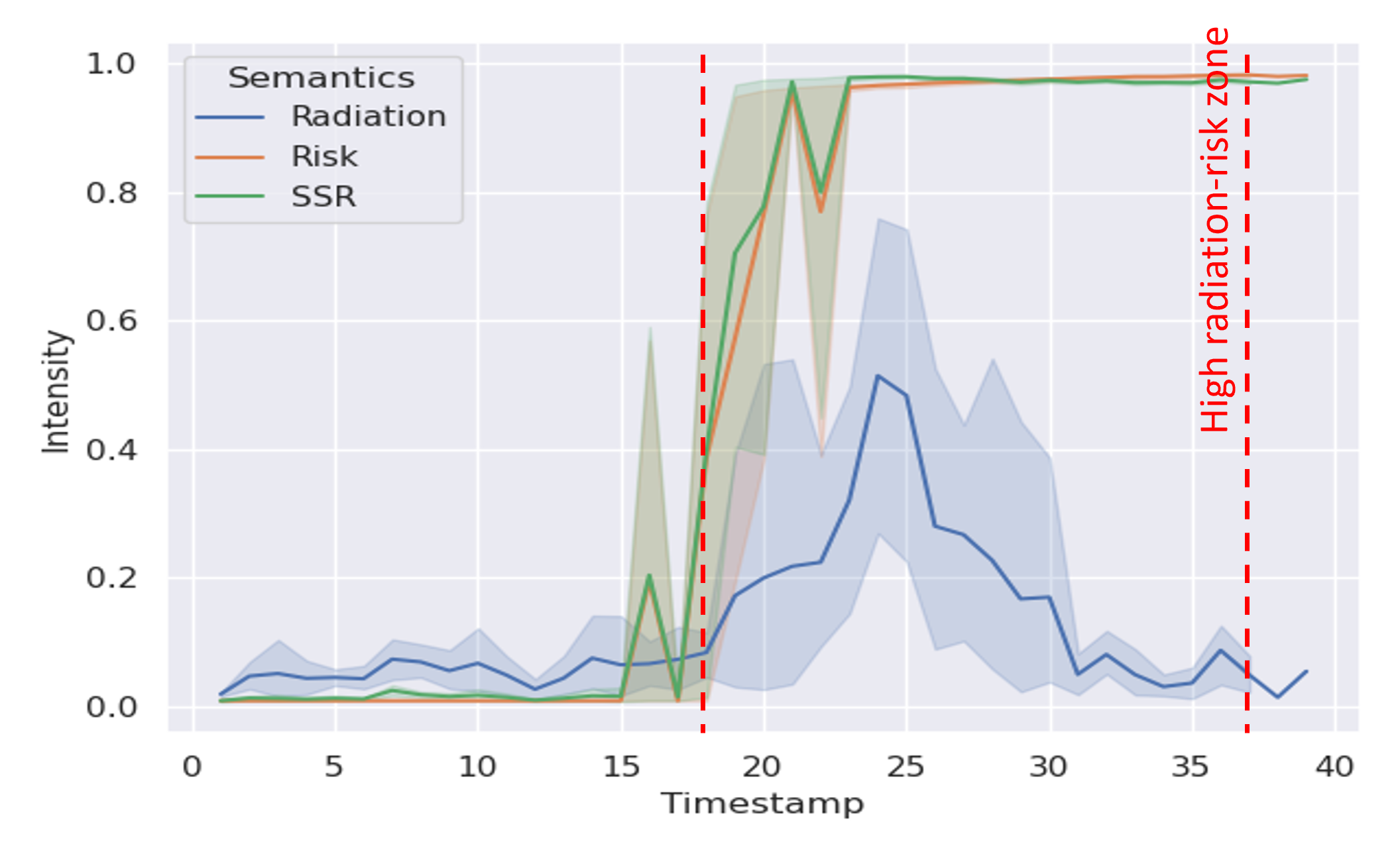}
        \includegraphics[width=0.48\linewidth]{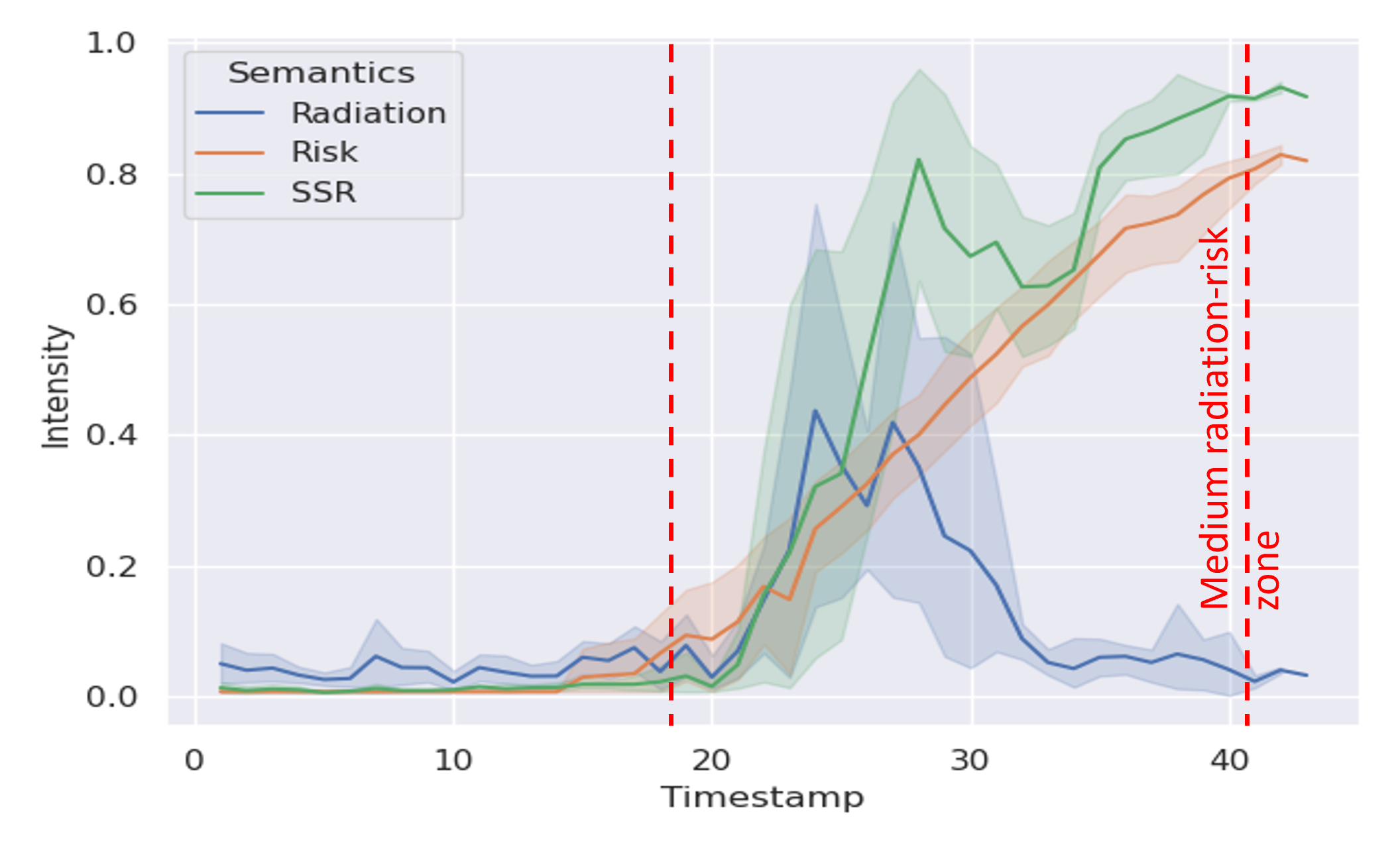}
    }
    
    \subfloat[Risk-noise combination]{
        \includegraphics[width=0.48\linewidth]{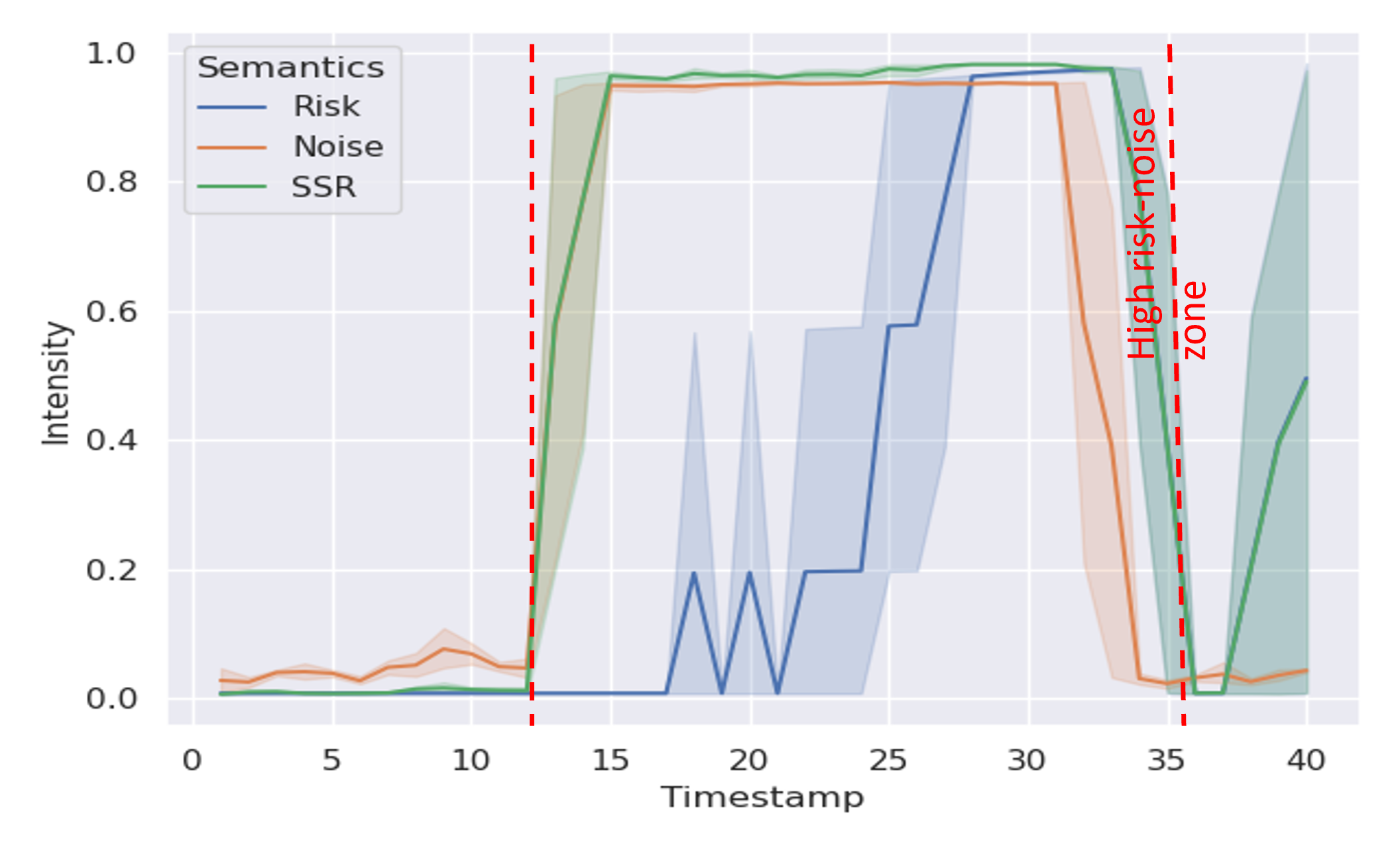}
        \includegraphics[width=0.48\linewidth]{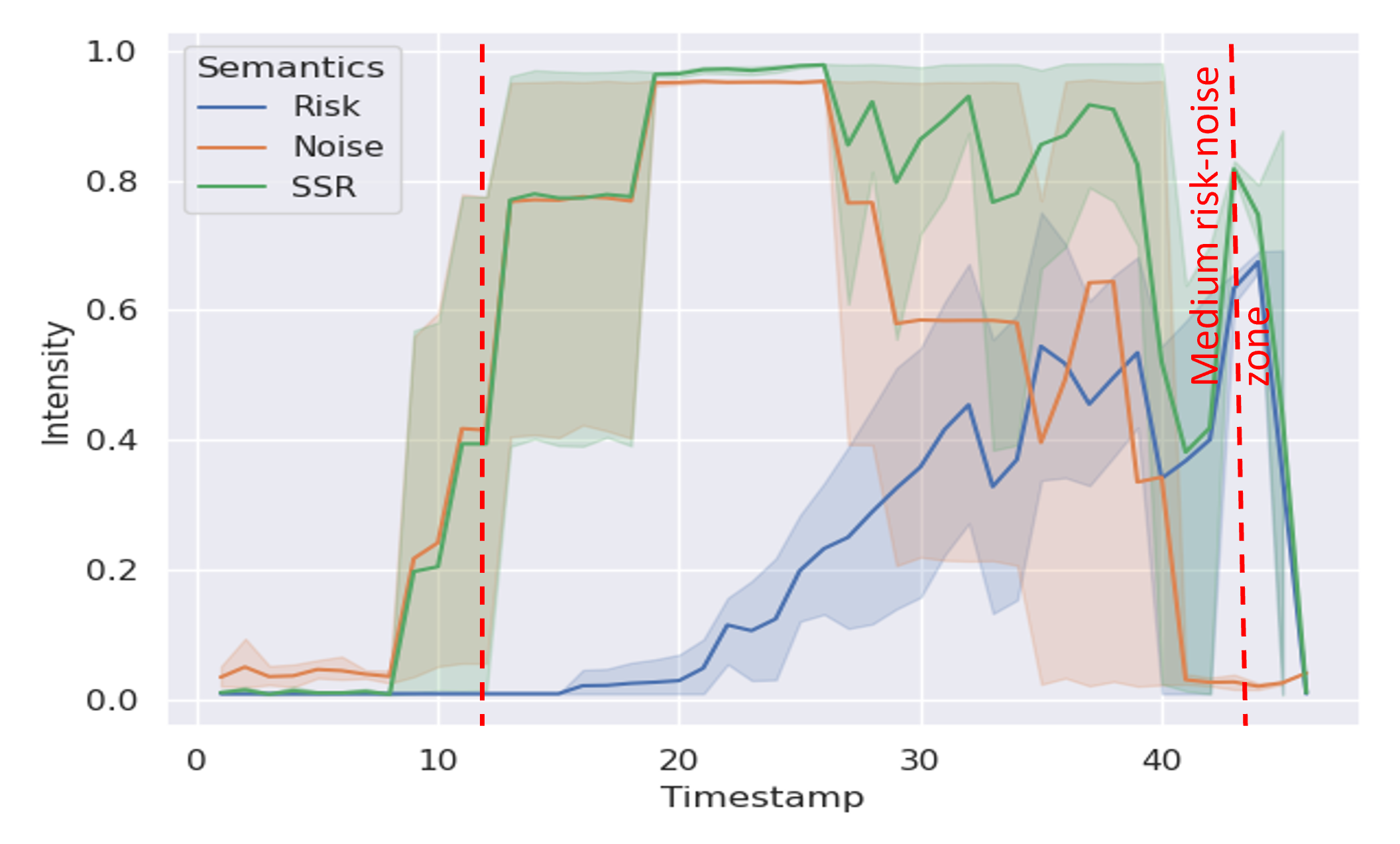}
    }
    
    \subfloat[SHA-noise combination]{
        \includegraphics[width=0.48\linewidth]{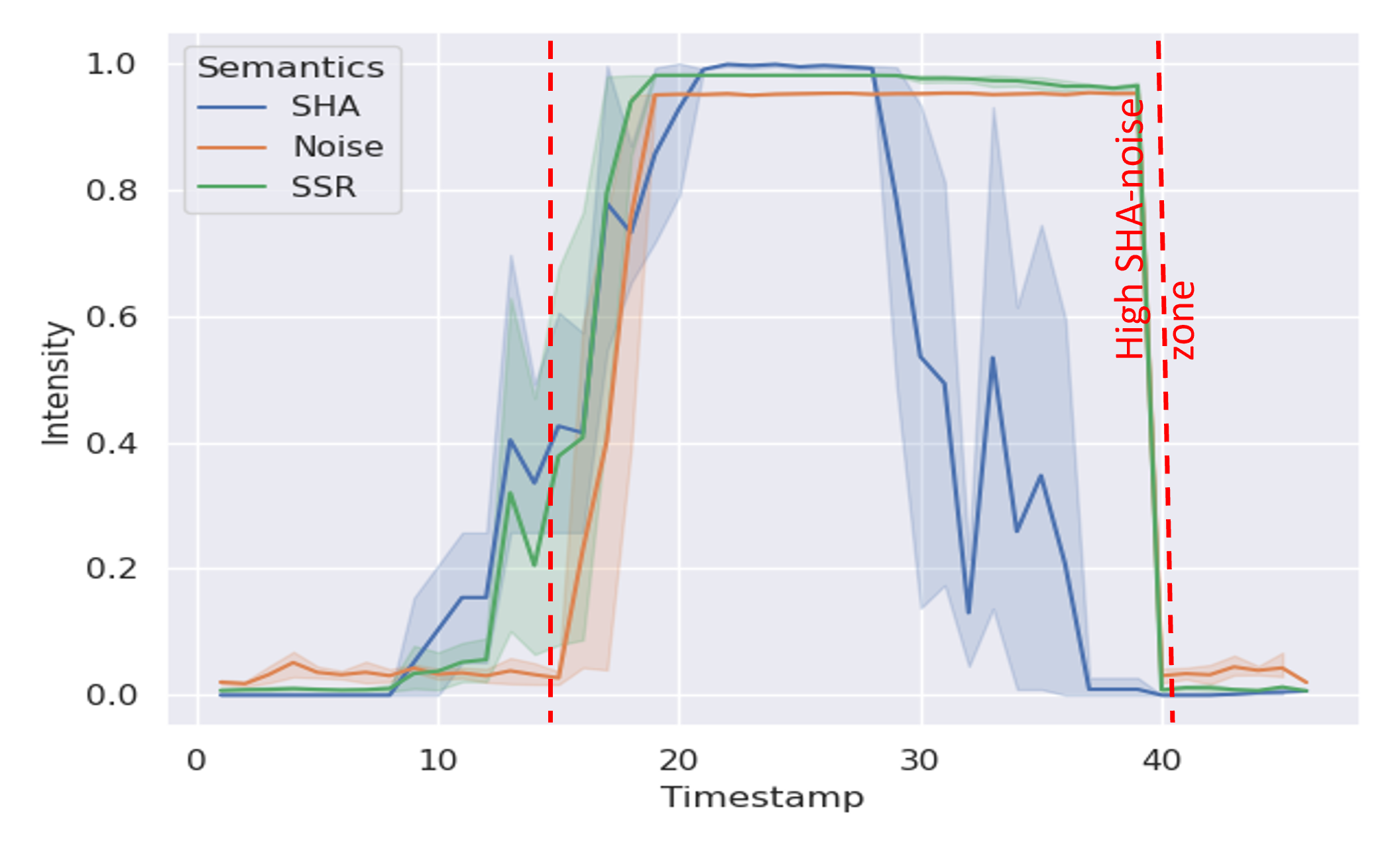}
        \includegraphics[width=0.48\linewidth]{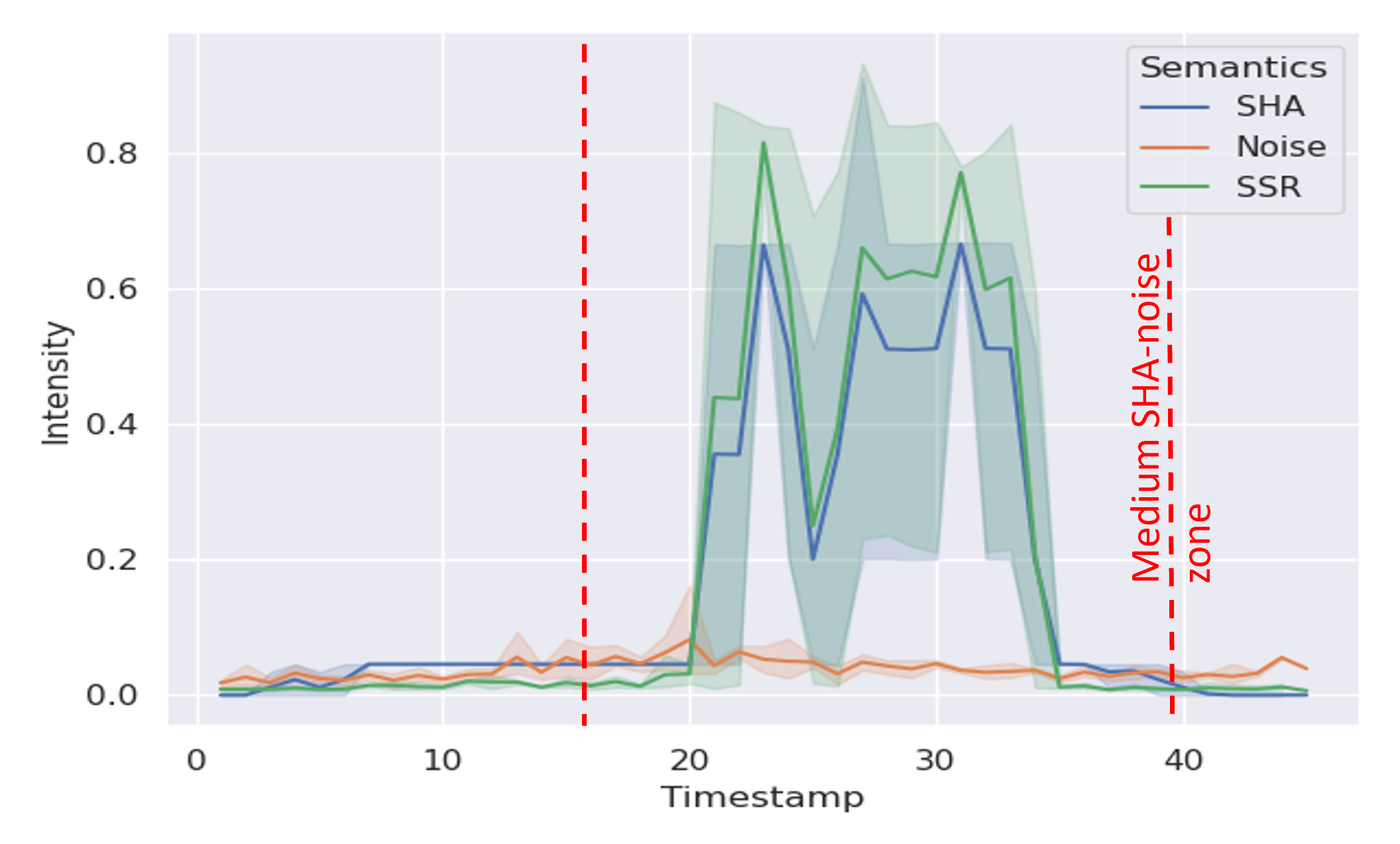}
    }

    \caption{SSR and environment semantics intensity timeline}
    \label{fig:timeline}
\end{figure*}

We collected data on all environment semantic indicators, SSR scores, and timestamps in Experiment II. We aligned the data in each scenario with the timestamp to reduce the error caused by mismatching. The processed results are shown in \cref{fig:timeline}.

Because of the limitation of the sensitivity analysis (unable to reflect the sensitivity in the complex system with multiple factors), we analyzed individual environment semantics correlations to the SSR score. Our framework is nonlinear. Hence we calculate Spearman's rank correlation coefficient ($\rho$) \cref{table: correlations} and Kendall's rank correlation coefficient ($\tau$) \cref{table: kendallcorrelations} to examine the relations among them. 

In \cref{table: correlations}, Spearman's rank correlation coefficient demonstrates that in most scenarios, environment semantics indicators show at least a ``weak" correlation (0.1 - 0.39) to the SSR score. In most scenarios with high-intensity semantics, the correlation index is above o.4 and can be considered a ``moderate" (0.4 - 0.69). If the semantics have high intensity with a long time, tables show  ``strong" (0.7 - 0.89) correlation \cite{schober2018correlation}. The table indicates that different semantics show at least moderate impacts on the SSR score, i.e. the SSR score can reflect the changes on the multiple semantics changes accordingly. If we connect to Experiment I, we find the results are consistent in the scope of responding to the situation changes correctly. It means our framework shows the generality ability when adding or removing semantics indicators.

Kendall's rank correlation coefficient \cite{chok2010pearson} is robust to outliers. Similarly, most results in \cref{table: kendallcorrelations} align the monotonic relationship (at least moderate positive correlation) shown in \cref{table: correlations} which is expected. However, Scenario 2 shows a "weak positive correlation" with noise. When we check the \cref{table: correlations} (a), the weak correlation is reasonable. We did not deploy artificial noise in that scenario. Hence the noise score is mainly affected by the robot's movement which stays at the bottom of the graph. It does not impact the changes in SSR score much as we expected. A similar situation happens in Scenario 8 and results in a weak correlation. In Scenario 11, there is no environment semantics showing a dominant impact on the SSR score from the \cref{table: correlations} (f). Hence, the correlation coefficients indicate weak or moderate correlations only.

Both Spearman's rank correlation coefficient and Kendall's rank correlation coefficient demonstrate that our framework reveals the situation changes and can adapt the complex scenarios with multiple semantics. These environment semantics shows considerable impacts on the SSR score that enable SSR as a trustworthy representative for warning situation changes.

\section{Discussion and Future Work}

Multi-robot deployments are expected in the future. SA is one of the prerequisites of prediction, planning, and decision-making. Our framework provides a way to obtain real-time SA that is intuitive and explainable to humans and easily usable from robots. From the scope of the experiments, the sensitivity analysis of Experiment I and the Spearman's, and Kendal's rank coefficient of Experiment II, along with the analysis of the figures, reveal that our framework is sensitive enough for individual semantics situations and responds correctly in semantics-rich situations.

Compared with deep learning methods, our framework is designed to apply expert knowledge instead of data-driven training, which avoids the issue of lacking datasets. It can be potentially explainable to a human, contrary to black box models. It enables humans to understand what exactly happens in the framework and makes it more intuitive for experts tuning the framework. Human operators and robots can obtain shared SA not only from the SSR score but also from the changes in individual semantics indicators, which enables them, to identify the exact situation onsite. This is crucial for real-world deployments in safety-critical applications. Also, context and semantics can be infinite. Robots may not be able to understand all situations with a complex combination of semantics. This metric can be used to make robots aware of whether they are in a semantics-rich situation beyond their capability to understand and whether they need help from human intelligence. Hence, both the semantics indicators and the aggregated metric SSR can be used in a prediction, planning, and decision-making framework, especially from Human-Robot Teams. 

Regarding the flexibility and generality of our proposed approach, Experiment II indicates that the framework is flexible enough to shift and comprise multiple environment semantics. It will remain robust, easy-explainable, and intuitive if more environment semantics are added or removed. Depending on the applications and mission, experts can directly adjust parameters to generalize the framework into a more reasonable representation of the given scenario. For instance, experts can highlight the weights of those semantics that are important to the goal of the mission, which makes the framework more sensitive and responsive to these semantics. The framework is not restricted to UGV deployments and can be adopted from different robotics platforms and required sensors, such as UAVs or heterogeneous multi-robot teams. 

We have some limitations also. The vision system constrains the implementation of the framework. Considering the training process, it is possible to fuse personal belonging detection and hazmat detection to simplify the deployment process. Or, we can adapt other state-of-the-art perception algorithms to improve the accuracy and real-time performance. However, our framework will scale nicely to continue being useful as the semantics capabilities of AI and computer vision continue to grow more powerful over time. 

\section{Conclusion}
In this paper, we proposed a semantics-based SA framework to represent and quantify the variety of semantic information and the overall information richness via the concepts of environment semantic indicators and the aggregated Situational Semantic Richness metric. We also provided an example implementation to process high-level environment semantic indicators that quantify the corresponding specific scope of the environment. Semantic perception capabilities of AI are still in an early stage of development. That's why we have chosen some relatively simple and robust examples in the experiments. However, experiments demonstrate our framework is capable of obtaining SA and indicate its extensibility to semantic-rich environments, and potential to involve multiple environment semantics. The modularized design increases the flexibility and it should adapt nicely as these AI capabilities grow.





\bibliographystyle{IEEEtran}

\begin{thebibliography}{10}
\providecommand{\url}[1]{#1}
\csname url@samestyle\endcsname
\providecommand{\newblock}{\relax}
\providecommand{\bibinfo}[2]{#2}
\providecommand{\BIBentrySTDinterwordspacing}{\spaceskip=0pt\relax}
\providecommand{\BIBentryALTinterwordstretchfactor}{4}
\providecommand{\BIBentryALTinterwordspacing}{\spaceskip=\fontdimen2\font plus
\BIBentryALTinterwordstretchfactor\fontdimen3\font minus \fontdimen4\font\relax}
\providecommand{\BIBforeignlanguage}[2]{{%
\expandafter\ifx\csname l@#1\endcsname\relax
\typeout{** WARNING: IEEEtran.bst: No hyphenation pattern has been}%
\typeout{** loaded for the language `#1'. Using the pattern for}%
\typeout{** the default language instead.}%
\else
\language=\csname l@#1\endcsname
\fi
#2}}
\providecommand{\BIBdecl}{\relax}
\BIBdecl

\bibitem{chiou2022robot}
M.~Chiou, G.-T. Epsimos, G.~Nikolaou, P.~Pappas, G.~Petousakis, S.~M{\"u}hl, and R.~Stolkin, ``Robot-assisted nuclear disaster response: Report and insights from a field exercise,'' in \emph{2022 IEEE/RSJ International Conference on Intelligent Robots and Systems (IROS)}.\hskip 1em plus 0.5em minus 0.4em\relax IEEE, 2022, pp. 4545--4552.

\bibitem{10018727}
T.~Ruan, H.~Wang, R.~Stolkin, and M.~Chiou, ``A taxonomy of semantic information in robot-assisted disaster response,'' in \emph{2022 IEEE International Symposium on Safety, Security, and Rescue Robotics (SSRR)}, 2022, pp. 285--292.

\bibitem{rustam2023status}
R.~Stolkin, N.~Molitor, P.~Berben, J.~Verbeek, T.~Reedman, H.~Burtin, L.-H. Velnom, P.~Garrec, J.~Van Den~Bosch, T.~Braunroth, M.~Knaack, A.~Wernke, S.~Kawatsuma, Y.~Yamamoto, I.~Szoke, I.~Dalyaev, A.~Lopota, E.~Nikitin, A.~Tribunsky, A.~Labba, P.~Picca, I.~Tellam, and R.~Reid, ``Status, barriers and cost-benefits of robotic and remote systems applications in nuclear decommissioning and radioactive waste management,'' no. NEA/RWM/R(2022)1, 2023.

\bibitem{FCN}
J.~Long, E.~Shelhamer, and T.~Darrell, ``Fully convolutional networks for semantic segmentation,'' in \emph{2015 IEEE Conference on Computer Vision and Pattern Recognition (CVPR)}, 2015, pp. 3431--3440.

\bibitem{3Dsemantic}
X.~Li, H.~Ao, R.~Belaroussi, and D.~Gruyer, ``Fast semi-dense 3d semantic mapping with monocular visual slam,'' in \emph{2017 IEEE 20th International Conference on Intelligent Transportation Systems (ITSC)}, 2017, pp. 385--390.

\bibitem{endsley2017toward}
M.~R. Endsley, ``Toward a theory of situation awareness in dynamic systems,'' in \emph{Situational awareness}.\hskip 1em plus 0.5em minus 0.4em\relax Routledge, 2017, pp. 9--42.

\bibitem{bavle2021slam}
H.~Bavle, J.~L. Sanchez~Lopez, E.~Schmidt~F, and H.~Voos, ``From slam to situational awareness: Challenges and survey,'' 2021.

\bibitem{stanton2017state}
N.~A. Stanton, P.~M. Salmon, G.~H. Walker, E.~Salas, and P.~A. Hancock, ``State-of-science: situation awareness in individuals, teams and systems,'' \emph{Ergonomics}, vol.~60, no.~4, pp. 449--466, 2017.

\bibitem{endsley1988situation}
M.~R. Endsley, ``Situation awareness global assessment technique (sagat),'' in \emph{Proceedings of the IEEE 1988 national aerospace and electronics conference}.\hskip 1em plus 0.5em minus 0.4em\relax IEEE, 1988, pp. 789--795.

\bibitem{taylor2017situational}
R.~M. Taylor, ``Situational awareness rating technique (sart): The development of a tool for aircrew systems design,'' in \emph{Situational awareness}.\hskip 1em plus 0.5em minus 0.4em\relax Routledge, 2017, pp. 111--128.

\bibitem{endsley1995measurement}
M.~R. Endsley, ``Measurement of situation awareness in dynamic systems,'' \emph{Human factors}, vol.~37, no.~1, pp. 65--84, 1995.

\bibitem{hooey2011modeling}
B.~L. Hooey, B.~F. Gore, C.~D. Wickens, S.~Scott-Nash, C.~Socash, E.~Salud, and D.~C. Foyle, ``Modeling pilot situation awareness,'' in \emph{Human modelling in assisted transportation: Models, tools and risk methods}.\hskip 1em plus 0.5em minus 0.4em\relax Springer, 2011, pp. 207--213.

\bibitem{mcaree2018quantifying}
O.~McAree, J.~Aitken, and S.~Veres, ``Quantifying situation awareness for small unmanned aircraft: Towards routine beyond visual line of sight operations,'' \emph{The Aeronautical Journal}, vol. 122, no. 1251, pp. 733--746, 2018.

\bibitem{dini2017measurement}
A.~Dini, C.~Murko, S.~Yahyanejad, U.~Augsd{\"o}rfer, M.~Hofbaur, and L.~Paletta, ``Measurement and prediction of situation awareness in human-robot interaction based on a framework of probabilistic attention,'' in \emph{2017 IEEE/RSJ International Conference on Intelligent Robots and Systems (IROS)}.\hskip 1em plus 0.5em minus 0.4em\relax IEEE, 2017, pp. 4354--4361.

\bibitem{huang2019ontology}
L.~Huang, H.~Liang, B.~Yu, B.~Li, and H.~Zhu, ``Ontology-based driving scene modeling, situation assessment and decision making for autonomous vehicles,'' in \emph{2019 4th Asia-Pacific Conference on Intelligent Robot Systems (ACIRS)}.\hskip 1em plus 0.5em minus 0.4em\relax IEEE, 2019, pp. 57--62.

\bibitem{tenorth2017representations}
M.~Tenorth and M.~Beetz, ``Representations for robot knowledge in the knowrob framework,'' \emph{Artificial Intelligence}, vol. 247, pp. 151--169, 2017.

\bibitem{armand2014ontology}
A.~Armand, D.~Filliat, and J.~Iba{\~n}ez-Guzman, ``Ontology-based context awareness for driving assistance systems,'' in \emph{2014 IEEE intelligent vehicles symposium proceedings}.\hskip 1em plus 0.5em minus 0.4em\relax IEEE, 2014, pp. 227--233.

\bibitem{shuang2014quantitative}
L.~Shuang, W.~Xiaoru, and Z.~Damin, ``A quantitative situational awareness model of pilot,'' in \emph{Proceedings of the International Symposium on Human Factors and Ergonomics in Health Care}, vol.~3, no.~1.\hskip 1em plus 0.5em minus 0.4em\relax SAGE Publications Sage CA: Los Angeles, CA, 2014, pp. 117--122.

\bibitem{nguyen2019review}
T.~Nguyen, C.~P. Lim, N.~D. Nguyen, L.~Gordon-Brown, and S.~Nahavandi, ``A review of situation awareness assessment approaches in aviation environments,'' \emph{IEEE Systems Journal}, vol.~13, no.~3, pp. 3590--3603, 2019.

\bibitem{gao2020uav}
X.~Gao, H.~Jia, Z.~Chen, G.~Yuan, and S.~Yang, ``Uav security situation awareness method based on semantic analysis,'' in \emph{2020 IEEE International Conference on Power, Intelligent Computing and Systems (ICPICS)}.\hskip 1em plus 0.5em minus 0.4em\relax IEEE, 2020, pp. 272--276.

\bibitem{ginesi2020autonomous}
M.~Ginesi, D.~Meli, A.~Roberti, N.~Sansonetto, and P.~Fiorini, ``Autonomous task planning and situation awareness in robotic surgery,'' in \emph{2020 IEEE/RSJ International Conference on Intelligent Robots and Systems (IROS)}.\hskip 1em plus 0.5em minus 0.4em\relax IEEE, 2020, pp. 3144--3150.

\bibitem{ghezala2014rsaw}
M.~W.~B. Ghezala, A.~Bouzeghoub, and C.~Leroux, ``Rsaw: A situation awareness system for autonomous robots,'' in \emph{2014 13th International Conference on Control Automation Robotics \& Vision (ICARCV)}.\hskip 1em plus 0.5em minus 0.4em\relax IEEE, 2014, pp. 450--455.

\bibitem{kruijff2012rescue}
G.-J.~M. Kruijff, F.~Pirri, M.~Gianni, P.~Papadakis, M.~Pizzoli, A.~Sinha, V.~Tretyakov, T.~Linder, E.~Pianese, S.~Corrao \emph{et~al.}, ``Rescue robots at earthquake-hit mirandola, italy: A field report,'' in \emph{2012 IEEE international symposium on safety, security, and rescue robotics (SSRR)}.\hskip 1em plus 0.5em minus 0.4em\relax IEEE, 2012, pp. 1--8.

\bibitem{nagatani2013emergency}
K.~Nagatani, S.~Kiribayashi, Y.~Okada, K.~Otake, K.~Yoshida, S.~Tadokoro, T.~Nishimura, T.~Yoshida, E.~Koyanagi, M.~Fukushima \emph{et~al.}, ``Emergency response to the nuclear accident at the fukushima daiichi nuclear power plants using mobile rescue robots,'' \emph{Journal of Field Robotics}, vol.~30, no.~1, pp. 44--63, 2013.

\bibitem{murphy2014disaster}
R.~R. Murphy, \emph{Disaster robotics}.\hskip 1em plus 0.5em minus 0.4em\relax MIT press, 2014.

\bibitem{reinmund2024variable}
T.~Reinmund, P.~Salvini, L.~Kunze, M.~Jirotka, and A.~F. Winfield, ``Variable autonomy through responsible robotics: Design guidelines and research agenda,'' \emph{ACM Transactions on Human-Robot Interaction}, vol.~13, no.~1, pp. 1--36, 2024.

\bibitem{methnani2024s}
L.~Methnani, M.~Chiou, V.~Dignum, and A.~Theodorou, ``Who’s in charge here? a survey on trustworthy ai in variable autonomy robotic systems,'' \emph{ACM Computing Surveys}, 2024.

\bibitem{chiou2021mixed}
M.~Chiou, N.~Hawes, and R.~Stolkin, ``Mixed-initiative variable autonomy for remotely operated mobile robots,'' \emph{ACM Transactions on Human-Robot Interaction (THRI)}, vol.~10, no.~4, pp. 1--34, 2021.

\bibitem{pappas2020vfh+}
P.~Pappas, M.~Chiou, G.-T. Epsimos, G.~Nikolaou, and R.~Stolkin, ``Vfh+ based shared control for remotely operated mobile robots,'' in \emph{2020 IEEE International Symposium on Safety, Security, and Rescue Robotics (SSRR)}.\hskip 1em plus 0.5em minus 0.4em\relax IEEE, 2020, pp. 366--373.

\bibitem{ramesh2022robot}
A.~Ramesh, R.~Stolkin, and M.~Chiou, ``Robot vitals and robot health: Towards systematically quantifying runtime performance degradation in robots under adverse conditions,'' \emph{IEEE Robotics and Automation Letters}, vol.~7, no.~4, pp. 10\,729--10\,736, 2022.

\bibitem{soltani2004fuzzy}
A.~Soltani and T.~Fernando, ``A fuzzy based multi-objective path planning of construction sites,'' \emph{Automation in construction}, vol.~13, no.~6, pp. 717--734, 2004.

\bibitem{voudoukis2017inverse}
N.~Voudoukis and S.~Oikonomidis, ``Inverse square law for light and radiation: A unifying educational approach,'' \emph{European Journal of Engineering and Technology Research}, vol.~2, no.~11, pp. 23--27, 2017.

\bibitem{yangvisual}
W.~Yang, X.~Wang, A.~Farhadi, A.~Gupta, and R.~Mottaghi, ``Visual semantic navigation using scene priors.''

\bibitem{nadaraya1964estimating}
E.~A. Nadaraya, ``On estimating regression,'' \emph{Theory of Probability \& Its Applications}, vol.~9, no.~1, pp. 141--142, 1964.

\bibitem{watson1964smooth}
G.~S. Watson, ``Smooth regression analysis,'' \emph{Sankhy{\=a}: The Indian Journal of Statistics, Series A}, pp. 359--372, 1964.

\bibitem{vaswani2017attention}
A.~Vaswani, N.~Shazeer, N.~Parmar, J.~Uszkoreit, L.~Jones, A.~N. Gomez, {\L}.~Kaiser, and I.~Polosukhin, ``Attention is all you need,'' in \emph{Advances in neural information processing systems}, 2017, pp. 5998--6008.

\bibitem{yolact-iccv2019}
D.~Bolya, C.~Zhou, F.~Xiao, and Y.~J. Lee, ``Yolact: {Real-time} instance segmentation,'' in \emph{ICCV}, 2019.

\bibitem{sobol2001global}
I.~M. Sobol, ``Global sensitivity indices for nonlinear mathematical models and their monte carlo estimates,'' \emph{Mathematics and computers in simulation}, vol.~55, no. 1-3, pp. 271--280, 2001.

\bibitem{RameshBraunRothfuss2023_1000159207}
A.~Ramesh, C.~A. Braun, T.~Ruan, S.~Rothfuß, S.~Hohmann, R.~Stolkin, and M.~Chiou, ``\BIBforeignlanguage{english}{Experimental evaluation of model predictive mixed-initiative variable autonomy systems applied to human-robot teams},'' in \emph{\BIBforeignlanguage{english}{2023 IEEE International Conference on Systems, Man, and Cybernetics (SMC), October 1-4, 2023, Hyatt Maui, Hawaii, USA}}.\hskip 1em plus 0.5em minus 0.4em\relax IEEE, 2023.

\bibitem{schober2018correlation}
P.~Schober, C.~Boer, and L.~A. Schwarte, ``Correlation coefficients: appropriate use and interpretation,'' \emph{Anesthesia \& analgesia}, vol. 126, no.~5, pp. 1763--1768, 2018.

\bibitem{chok2010pearson}
N.~S. Chok, ``Pearson's versus spearman's and kendall's correlation coefficients for continuous data,'' Ph.D. dissertation, University of Pittsburgh, 2010.

\end{thebibliography}

\addtolength{\textheight}{-12cm}  

\end{document}